\documentclass[conference, 9pt]{IEEEtran}
\pdfoutput=1
\IEEEoverridecommandlockouts
\usepackage{cite}
\usepackage{amsmath,amssymb,amsfonts}
\usepackage{algorithmic}
\usepackage{graphicx}
\usepackage{multirow}
\usepackage{textcomp}
\usepackage[table,xcdraw]{xcolor}

\usepackage{textcomp}
\usepackage[ruled,vlined]{algorithm2e}

\usepackage{xspace}
\usepackage{subfigure}
\usepackage{graphicx}
 \usepackage{enumitem}
\usepackage{tablefootnote}
\usepackage{url}

\usepackage{xcolor}
\def\BibTeX{{\rm B\kern-.05em{\sc i\kern-.025em b}\kern-.08em
    T\kern-.1667em\lower.7ex\hbox{E}\kern-.125emX}}
    
\newcommand\ours{{\textit{Collate}}\xspace}
\begin{document}


\title{Collate: Collaborative Neural Network Learning for Latency-Critical Edge Systems

\thanks{\copyright~2022 IEEE. Personal use of this material is permitted. Permission from IEEE must be obtained for all other uses, in any current or future media, including reprinting/republishing this material for advertising or promotional purposes, creating new collective works, for resale or redistribution to servers or lists, or reuse of any copyrighted component of this work in other works. This is the author's accepted version of the article published in the 2022 IEEE 40th International Conference on Computer Design (ICCD), pp. 627-634, 2022, DOI: 10.1109/ICCD56317.2022.00097.

This study is supported under the RIE2020 Industry Alignment Fund – Industry Collaboration Projects (IAF-ICP) Funding Initiative, as well as cash and in-kind contribution from the industry partner, HP Inc., through the HP-NTU Digital Manufacturing Corporate Lab (I1801E0028).}
}


\author{\IEEEauthorblockN{Shuo Huai\IEEEauthorrefmark{3}\IEEEauthorrefmark{4}, Di Liu\IEEEauthorrefmark{4}, Hao Kong\IEEEauthorrefmark{3}\IEEEauthorrefmark{4}, Xiangzhong Luo\IEEEauthorrefmark{3}, Weichen Liu\IEEEauthorrefmark{3}, \\ Ravi Subramaniam\IEEEauthorrefmark{2}, Christian Makaya\IEEEauthorrefmark{2} and Qian  Lin\IEEEauthorrefmark{2}}
\IEEEauthorblockA{
\IEEEauthorrefmark{3}School of Computer Science and Engineering, Nanyang Technological University, Singapore\\
\IEEEauthorrefmark{4}HP-NTU Digital Manufacturing Corporate Lab, Nanyang Technological University, Singapore\\
\IEEEauthorrefmark{2}HP Inc., Palo Alto, California, USA \\
Email: \{shuo001, liu.di, kong.hao, xiangzho001, liu\}@ntu.edu.sg, \{ravi.subramaniam, christian.makaya, qian.lin\}@hp.com
}
}

\maketitle

\begin{abstract}

Federated Learning (FL) empowers multiple clients to collaboratively learn a model, enlarging the training data of each client for high accuracy while protecting data privacy. However, when deploying FL in real-time edge systems, the heterogeneity of devices among systems has a severe impact on the performance of the inferred model. Existing optimizations on FL focus on improving the training efficiency but fail to speed up inference, especially when there is a latency constraint. In this work, we propose \ours, a novel training framework that collaboratively learns heterogeneous models to meet the latency constraints of multiple edge systems simultaneously. We design a dynamic zeroizing-recovering method to adjust each local model architecture for high accuracy under its latency constraint. A proto-corrected federated aggregation scheme is also introduced to aggregate all heterogeneous local models, satisfying the latency constraint of different systems with only one training process and maintaining high accuracy. Extensive experiments indicate that, compared to state-of-the-art methods and under a latency constraint, our extended models can improve the accuracy by 1.96\% on average, and our shrunk models can also obtain a 3.09\% accuracy improvement on average, with almost no extra training overhead. The related codes and data will be available at
\end{abstract}

\begin{IEEEkeywords}
edge devices, edge intelligence, neural network learning, federated learning, inference efficiency
\end{IEEEkeywords}

\section{Introduction}

Deep Neural Networks (DNNs) have brought significant breakthroughs in many different applications, such as image recognition and natural language processing \cite{DLSurvey}. With the emphasis on data privacy and concerns over transmission stability, current DNN applications are increasingly deployed on edge devices, such as autonomous vehicles, healthcare devices, etc \cite{DLSurvey}. Meanwhile, DNN models require a huge amount of training data to improve accuracy \cite{zhu2016we}, but data in most industries are protected by privacy laws and thus are required to be in the form of isolated islands.



\begin{figure}[!t]
    \centering
    \includegraphics[width=0.48\textwidth]{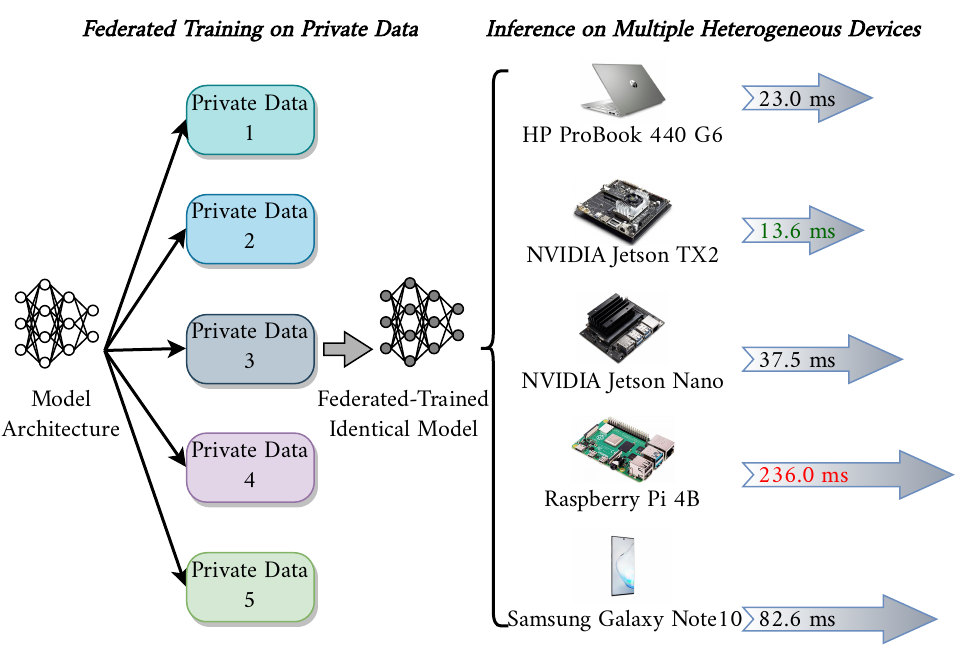}
    \caption{Latency of the federated-trained AlexNet on various edge systems.}
    \label{fig:motivation}
\end{figure}

Federated Learning (FL) is designed to coordinate multiple clients to train a DNN model collaboratively without sharing their original local data, and it is capable of preserving the data privacy and achieving better accuracy than each individual client training with only its local data \cite{yang2019federated}. This addresses the issue of insufficient individual training data. However, in addition to accuracy, latency is also an important metric for edge intelligent systems \cite{wang2019high}. With diverse edge devices that demonstrate different computational capabilities emerging, the identical model trained by FL is not efficient for all participating edge systems and cannot achieve a good balance between latency and accuracy.  As demonstrated in Fig.~\ref{fig:motivation}, when deploying the model into various edge systems, the latency differs from $13.6$ ms to $236.0$ ms. When this application has a latency constraint (e.g., $30.0$ ms), some systems can infer more complex models for higher accuracy, whereas others cannot even deploy the model due to the limited memory and computational resources.


Take a practical scenario from WeBank as an example \cite{cheng2020federated}. They need an edge FL framework in which surveillance video data collected and stored in the edge cloud of each surveillance company are not required to be uploaded to a central cloud for centralized model training. After each local training iteration, only the model parameters from each surveillance company are sent to the FL server for aggregation. The final federated-trained model is distributed to the participating surveillance companies for object detection. However, different companies can use distinct edge devices and this application features real-time, so it is necessary to train heterogeneous models for different companies to meet their latency constraints.





Some efforts have been made to use different DNN architectures to fit various clients in the training stage, known as Heterogeneous FL \cite{zhang2021parameterized, xu2021helios,diao2020heterofl,li2019fedmd,tan2022fedproto}. These methods mainly fall into two categories: one is to fine-tune models for different clients from an identical global model, and the other is to directly learn from heterogeneous models without the same global model. However, these existing approaches are designed for accelerating the training stage and cannot optimize the inference latency directly. When the first one is employed in a latency-critical system, it can only train a specific network for one system at a time until it trains for all systems using multiple FL processes, which imposes enormous training overhead. Although the second one can directly provide heterogeneous model architectures for different systems,  it requires some public datasets for transfer learning \cite{li2019fedmd} or an extra dataset for prototype learning \cite{tan2022fedproto}. Moreover, the absence of the same global model lead to an accuracy drop of up to  $10\%$  \cite{yuan2019distributed}. To guarantee the generality and accuracy of the FL scheme, our method should be based on the first one.




Meanwhile, when optimizing models under the latency constraint, we should not only reduce latency for low-end systems but also extend models for powerful systems (e.g., Jetson TX2 in Fig.~\ref{fig:motivation}) to improve their accuracy \cite{tan2019efficientnet}, in contrast to existing heterogeneous FL that only reduces the training cost by shrinking models. Thus, we integrate the model extension into our learning framework to better utilize each client. To our best knowledge, this is the first paper to optimize FL to simultaneously meet the latency constraints of all participating systems while obtaining high accuracy. Specifically, our main contributions are summarized as:

\begin{itemize}
\item We propose a novel model learning framework, \ours, that cultivates optimal DNN architectures collaboratively for multiple edge systems to obtain higher accuracy and satisfy their latency constraints with only one training process.

\item We present a proto-corrected aggregation scheme in the global training process to effectively aggregate all heterogeneous models from each edge system for higher accuracy.

\item We design a latency-aware local training scheme by a dynamic zeroizing-recovering training process. It extends the exploration space of \ours to discover the optimal DNN architecture for each edge system. 



\item We demonstrate the effectiveness of \ours with extensive experiments. Compared to the state-of-the-art methods and under the same latency constraints, our extended models can improve the accuracy by $1.96\%$  on average, and the accuracy of shrunk models outperforms others by $3.09\%$ on average.



\end{itemize}


\begin{figure*}[!t]
    \centering
    \includegraphics[width=0.98\textwidth]{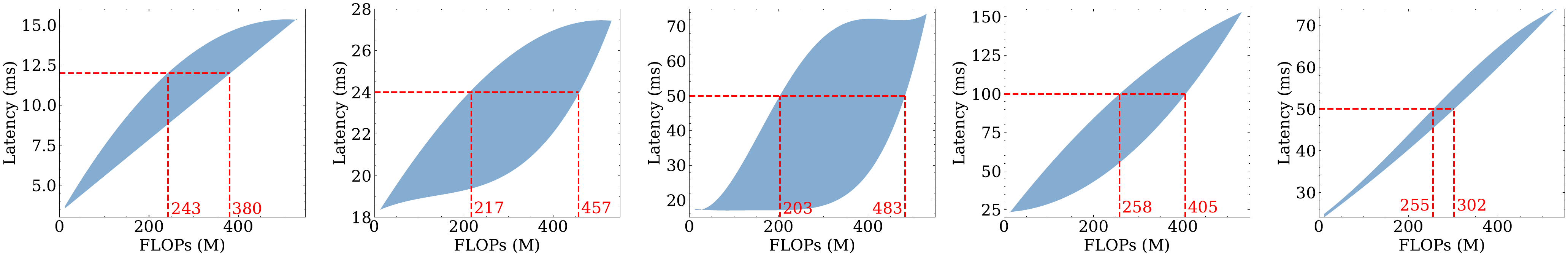}
    \caption{Relationship between latency and FLOPs of ResNet variants on devices (From left to right are HP Probook, Jetson TX2, Jetson Nano, Pi and Note10).}
    \label{fig:latency-flop}
\end{figure*}

\begin{figure*}[!t]
    \centering
    \includegraphics[width=0.98\textwidth]{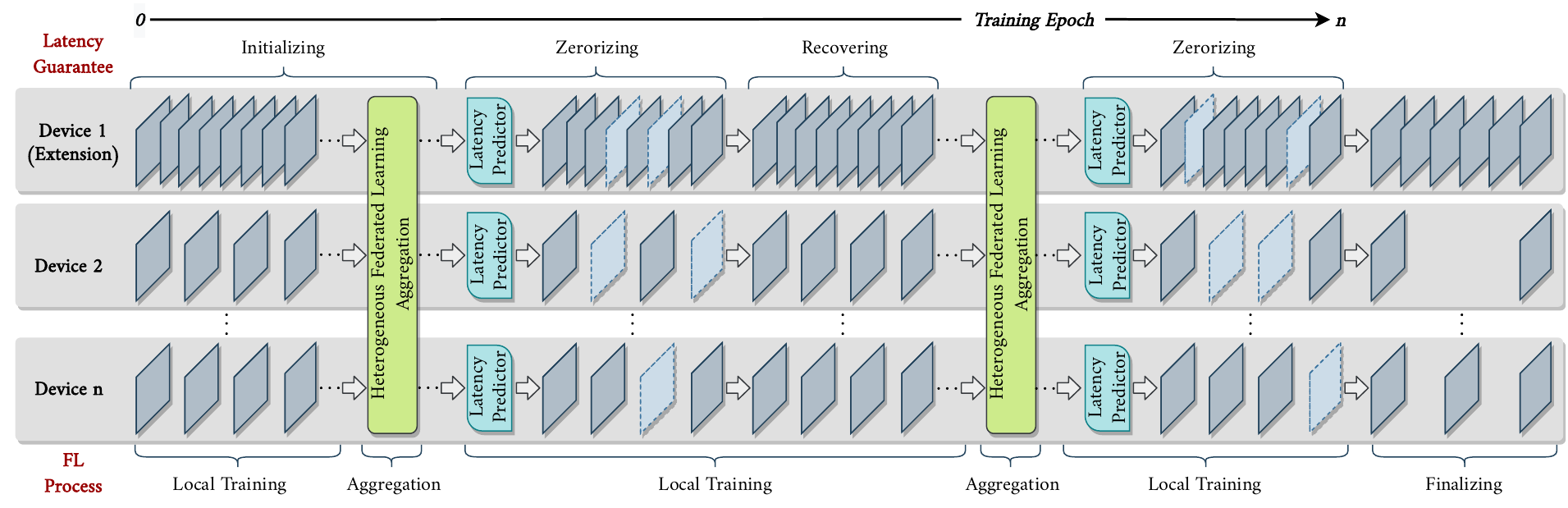}
    \caption{The process of \ours. The top depicts the latency guarantee component, while the bottom shows the heterogeneous FL component. }
    \label{fig:overview}
\end{figure*}

\section{Background \& Related Work}
\label{section:relatedwork}

In this section, we introduce the preliminaries of FL, heterogeneous FL and latency prediction, including the advantages and disadvantages of some related work.
\subsection{Federated Learning}

In the traditional FL algorithm (FedAvg \cite{mcmahan2017communication}), multiple edge systems collaborate to train the same model on their respective local training data to solve issues of data isolated islands and protect privacy. During this training step, different systems transmit model parameters to the server after every $le$ local training epochs. Then the server executes model aggregation. Considering this FL process includes $n$ systems, the loss function for system $i$ is $f_i$. Let $w$ represent the model parameter, and then the training optimization goal becomes:
\begin{equation}
\mathop{min}\limits_{w} f(w):=\frac{1}{n}\sum_{i=1}^nf_i(w)
\end{equation}
And the training process can be formulated as:
\begin{equation}
w_i^{t+(e+1)} = w_i^{t+e} - \eta_i^{t+e} \nabla f_i(w_i^{t+e})
\end{equation}
\begin{equation}
\label{eq-fl-1}
w^{t+1} = \frac{1}{n}\sum_{i=1}^n w_i^{t+le}
\end{equation}
where $w_i^{t+e}$ denotes the local model parameter on system $i$ after $t$ communication rounds and $e$ local training steps. $\eta$ is the learning rate, and $\nabla$ represents the derivative. From Eq.~(\ref{eq-fl-1}), we can see that the traditional FL \cite{mcmahan2017communication} is based on the fact that all local model architectures of different systems are the same for aggregation. 

\subsection{Heterogeneous Federated Learning}

Two major challenges involved in the traditional FL process are statistical heterogeneity and hardware heterogeneity among different systems \cite{li2020federated}. Statistical heterogeneity means the data on different systems are not independent and identically distributed (Non-IID) \cite{zhao2018federated}, and different systems have different information to learn, while hardware heterogeneity refers to various devices participating in the FL process having different computational abilities. Different heterogeneous FL methods are proposed for these two challenges.


\textbf{Statistical Heterogeneity:}
Most existing works on statistical heterogeneity aim to achieve higher accuracy on each local data. They use different or identical model architectures with different parameters for each participating system by customizing the global model. 
Wang \textit{\textit{et al.} } \cite{wang2019federated} proposed to fine-tune some or all parameters of a trained global model using each client's local data. 
Jiang \textit{et al.}  \cite{jiang2019improving} 
proposed to combine meta-learning with FL for customizing models to recognize each local data pattern. Khodak \textit{et al.}  \cite{khodak2019adaptive} proposed to improve the accuracy of statistically heterogeneous FL via online convex optimization theory and meta-learning. 
%
%
These methods focus on improving the accuracy of local models on the local dataset of each system. They are applied to the scenario where most of the inference data of each system match the pattern of its corresponding training set. This paper aims to learn different local models for all participating edge systems to meet their latency requirements. Thus, when considering statistical heterogeneity, these aforementioned methods can be used as the back-end of our method to achieve higher accuracy.

\textbf{Hardware Heterogeneity:}
Current research on hardware heterogeneity concentrates on simplifying the model architecture to accelerate training and reduce the model size on devices with low computational ability. Caldas  \textit{et al.}  \cite{caldas2018expanding} proposed to randomly select small subsets of the global model with the expected model volume to provide a reduction in both client-to-server communication and local computation during the training.  Li  \textit{et al.}  \cite{li2021hermes} proposed Hermes to find small sub-networks for some devices by applying the structured pruning. Jeong  \textit{et al.}  \cite{jeong2018communication} proposed to use knowledge distillation to reduce model size in some low-end devices. Xu  \textit{et al.}  \cite{xu2021helios} proposed Helios to accelerate the devices with weak computational capacities by dynamically compressing the global model into an expected volume during the training stage.  Diao  \textit{et al.}  \cite{diao2020heterofl} proposed HeteroFL to assign models with different computational complexities to each device in advance according to the computational ability of each device. Also, they proposed a method to aggregate heterogeneous local models to produce a global model during the training stage.

These methods are proposed to reduce the training time or the communication cost when applying on-device training.
However, FL is not just on-device training \cite{wibowo2021mobile}. And typically, the usage (inference) of a model lasts longer than its generation (training). Thus, inference optimization is more critical for the FL framework. Although some of these methods can be modified to accelerate inference on low-end devices, these methods designed for the global model lead to a large accuracy loss for each local model. Meanwhile, these methods did not consider improving accuracy for powerful devices by extending models, and their aggregation methods introduce an accuracy drop to large models, as illustrated in Section \ref{section:experiments}.





\subsection{Latency Prediction}

Optimizing models based on latency can better explore hardware characteristics, providing additional advantages in the trade-off between accuracy and latency \cite{yang2018netadapt}. To obtain high accuracy under latency constraints, we dynamically adjust the local model architectures during their local training, and the new model architecture needs to satisfy the latency constraint. However, measuring latency on a device will interrupt the training process and usually takes minutes \cite{ChamNet}, especially for off-device training \cite{wibowo2021mobile}.
And as shown in Fig.~\ref{fig:latency-flop},  the relationship between latency and floating-point operations (FLOPs) of a model is weak, making it unlikely to calculate the latency with an arithmetic function with respect to the FLOPs. A latency predictor is necessary for latency-critical systems.


As the overhead of on-device measurement is huge, previous works for latency-critical systems use hardware simulator-based \cite{gholami2018squeezenext}, look-up table (LUT)-based \cite{ChamNet, yang2018netadapt, FBNet} or neural network-based \cite{justus2018predicting, huai2021zerobn} predictors to estimate the latency of DNN on a certain device. Most commercial devices are black boxes to users, so it is difficult to emulate hardware by analyzing its resources and scheduling algorithms for building hardware simulators. On the other hand, LUT-based predictors calculate the latency of DNN by summing up the recorded latency of each layer, resulting in that it only suiting models made up of pre-defined layer structures. During our latency-critical learning process, the number of input channels and kernels, and other size-related parameters of each layer will be updated continuously to find the best model architecture under the latency constraint. It is impractical for LUT to preserve the whole design space. Thus, in this work, we follow \cite{huai2021zerobn} to build a hardware-customized latency predictor based on the Backpropagation (BP) \cite{bp} neural network.




\section{Methodology}
\label{section:methology}

In this section, we demonstrate our model learning framework, which collaboratively trains heterogeneous DNNs for different edge devices to meet their respective latency constraints simultaneously,  while optimizing the model architecture for high accuracy. As shown in Fig.~\ref{fig:overview}, this framework mainly includes two parts: the latency guarantee for each participant and the heterogeneous federated learning process. We also employ model extension for powerful clients, and we use the latency predictor to optimize this flow. Fig.~\ref{fig:overview} also illustrates the change of local models on each device during our training process, where we use kernels in a layer as an example to show this training process, and each oblique rectangle represents a kernel in this layer. The dashed line means this kernel is shielded in the current epoch. In the following, we first briefly introduce the hardware-customized latency predictor and the model extension. Then we present the latency guarantee, which is integrated into the local training of the FL process with almost no extra training overhead.  Finally, we introduce our heterogeneous FL, mainly focusing on the heterogeneous aggregation algorithm.

\subsection{Latency Predictor \& Model Extension}
\textbf{Latency Predictor:} Based on \cite{huai2021zerobn}, we build a hardware-customized latency predictor using a three-layer BP network. The BP network is a lightweight neural network, and its forward propagation is shown in Eq.~(\ref{bp}), where $x$ is the vector of inputs, $W^L$ is the weight matrix, $L$ is the index of layers, and $\sigma^L$ is the activation function at layer $L$. Thus, hundreds of multiplication and addition operations are enough to obtain the latency with the BP network. Also, the updating scheme of the BP network is simple, so its training is not time-consuming.
\begin{equation}
\label{bp}
   g(x) = \sigma^2(W^2\sigma^{1}(W^{1} \sigma^0(W^0x))) 
\end{equation}

We first build a pool of models, which comprises thousands of single-layer models with various operations like Convolution and Pooling, configured with different configurations, to train the latency predictor. The latency of single-layer models is only a few milliseconds, so their measurement does not take much time. After training, by feeding the configuration of a layer, the latency predictor can predict its latency. For a real model, adding the latency of all layers in the latency-weighted longest path can get the whole model latency. The latency predictor is trained in advance for each device and is used in model learning. Assessing the latency predictor of Jetson TX2 with real models shows its variation is about 6.12\%.



\begin{figure}[!t]
    \centering
    \includegraphics[width=0.48\textwidth]{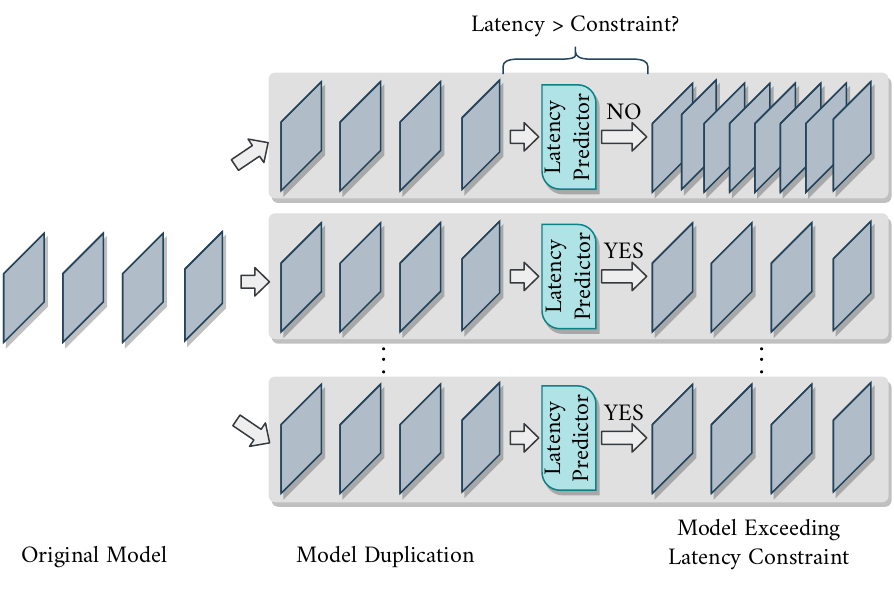}
    \caption{The model Extension scheme of \ours.}
    \label{fig:preprocess}
\end{figure}


\textbf{Model Extension:} 
In this work, we directly optimize models under the inference latency constraint, so \ours not only reduces latency for low-end systems but also extends models for powerful systems to improve accuracy. As illustrated in Fig.~\ref{fig:preprocess}, when the latency of the original model is lower than the latency constraint, we can uniformly extend the model \cite{gordon2018morphnet} on this device to obtain a larger model. The uniform extension means all layers are extended with a constant width multiplier, which enables the new model to exceed the system latency constraint. Then our following learning method can ensure the latency constraint is satisfied again.  The model extension improves the accuracy for powerful devices (see Section \ref{section:experiments}), but it introduces heterogeneity among different local models.

\subsection{Latency Constraint Guarantee -- Local Training}
\label{subsection:localtraining}

As shown in Fig \ref{fig:overview},  the latency guarantee flow is integrated into the local training of the FL process. This flow can be divided into three phases: Initializing, Zeroizing, and Recovering.  Algorithm~\ref{code:localtraining} demonstrates the details of local training with the latency constraint guarantee.


We first train the model for some communication rounds, called Initializing process, to let model parameters have informative values to reflect their contributions (\textit{Line 2-4}) instead of randomly initialized numbers. The proto-training will be introduced in the next section. We design a trainable mask layer following each convolution layer, and each value $\delta$ in the mask layer is multiplied into the output of each kernel in its previous convolution layer. Thus, the contribution metric we define is that kernels with larger absolute mask values provide more impact on the final accuracy. And zeroizing the mask value means removing the corresponding kernel from the previous convolution layer. Note that mask values can be fused into the weights of kernels before inference to eliminate their overhead. In the Zeroizing process, we first sort all contribution factors $\delta$ according to their absolute values to get the global contribution rank of kernels in the entire local model (\textit{Line 8}). During the whole training process, these contribution factors are jointly optimized with the network weights, hence the training scheme can automatically identify the contribution of each kernel. Next, we calculate the contribution factor threshold (\textit{Line 10}) by the latency predictor (\textit{Line 9}) and the contribution rank of the entire model. Finally, we zeroize those $\delta_i$ corresponding to the unimportant parts with these thresholds (\textit{Line 11}) to temporarily remove these weights for the current epoch. After this Zeroizing process, we extract a compact model that meets the latency constraint.

\begin{algorithm}[t]
\label{code:localtraining}

\KwIn{model, training settings, current\_round: $c$, start\_zero round: $s$, local\_epoch: $le$, latency constraint: $l$, prototype array: $pt$.}
\KwOut{updated local model, output prototype}

\caption{Latency Guarantee -- LocalTrain}
\begin{algorithmic}[1] 
\STATE Load private training data;\
\IF{$c < s$}
 \STATE $train\_with\_proto(pt)$ for $le$ epochs;\
\ELSE
\FOR{$e \leftarrow 1$ \textbf{to} $le$}
 \STATE $train\_with\_proto(pt)$;\
 \IF{$e \% 2\ != 0$ or $e$ is the last epoch}
 \STATE $Con\_{rank} \leftarrow $ Sort kernels by $|\delta|$;\
 \STATE $Shrink\_{ratio} \leftarrow$ Predictor($l,Con\_{rank}$);\
 
 \STATE Find threshold of $\delta$ by $Con\_{rank}$, $Shrink\_{ratio}$;\
 \STATE Zeroize $\delta_i$ if $\delta_i$ $<$ threshold;\
 \ENDIF
\ENDFOR
\ENDIF
\end{algorithmic}
\end{algorithm}

Following each Zeroizing process, we propose a Recovering process (\textit{Line 7}) to allow those zeroized parameters (i.e., $\delta_i$) to recover instead of being permanently removed. If any weights that are eliminated in the previous Zeroizing process but are potentially essential, the Recovering process can help them escape from zero and play a crucial part in the subsequent training process, the aggregation process, and even the inference processes. The Recovering process gives our method a chance to learn more efficient architecture with higher accuracy. Eq. (\ref{eq-zerobn-1}) indicates the calculation formula in a convolution layer with $\delta$ and activation function, where $L$ is the index of layers, $A^L$ is the output of layer $L$, $W^L$ is the weight between layer $L-1$ and $L$, and $\sigma^L$ is the activation function at layer $L$. $\circledast$ denotes the convolution operation.
\begin{equation}
\label{eq-zerobn-1}
 A^L = \sigma^L(\delta^{L} \cdot ( A^{L-1} \circledast W^L))
 \end{equation}
In the Recovering processing, we mainly focus on the updating of $\delta^{L}$ that are zeroized during the Zeroizing process. Eq. (\ref{eq-zerobn-2}) shows the gradients of $\delta^{L}$ to the final loss function.
\begin{equation}
\label{eq-zerobn-2}
   \frac{\partial loss}{\partial \delta^L } = \frac{\partial loss}{\partial A^L}\cdot {\sigma^L}^{'} \cdot ( A^{L-1} \circledast W^L)
 \end{equation}

Since we have zeroized some values in $\delta^{L}$ and the most often used activation function is \textit{ReLU}, the corresponding output of $\sigma^L$ is likewise zero under \textit{ReLU}. Although \textit{ReLU} is not differentiable at zero, it is widely accepted that its derivative is also zero. Thus, we can easily derive Eq. (\ref{eq-zerobn-4}).
\begin{equation}
\label{eq-zerobn-4}
  \delta^L = 0 \rightarrow {\sigma^L}^{'} = 0 \rightarrow \frac{\partial loss}{\partial \delta^L } = 0
\end{equation}
However, most training algorithms employ momentum in the optimizer to speed up convergence, as the momentum accumulates the gradients of the past steps to determine the direction to go in, rather than using only the gradient of the current step \cite{mm}. Eq. (\ref{eq-zerobn-5})-(\ref{eq-zerobn-6}) illustrates the updating rules of weight with momentum, in which $\eta$ is the learning rate, $z_k$ is the updated value of the last step, and $m$ is the accumulation coefficient. Thus, $w_k$ are easily updated to $w_{k+1}$ by combining current and past gradients.
\begin{equation}
\label{eq-zerobn-5}
    z_{k+1} = m\cdot z_k +  \frac{\partial loss}{\partial w_k}
\end{equation}
\begin{equation}
\label{eq-zerobn-6}
    w_{k+1} = w_k - \eta \cdot z_{k+1}
\end{equation}

\begin{figure}[t]
    \centering
    \includegraphics[width=0.49\textwidth]{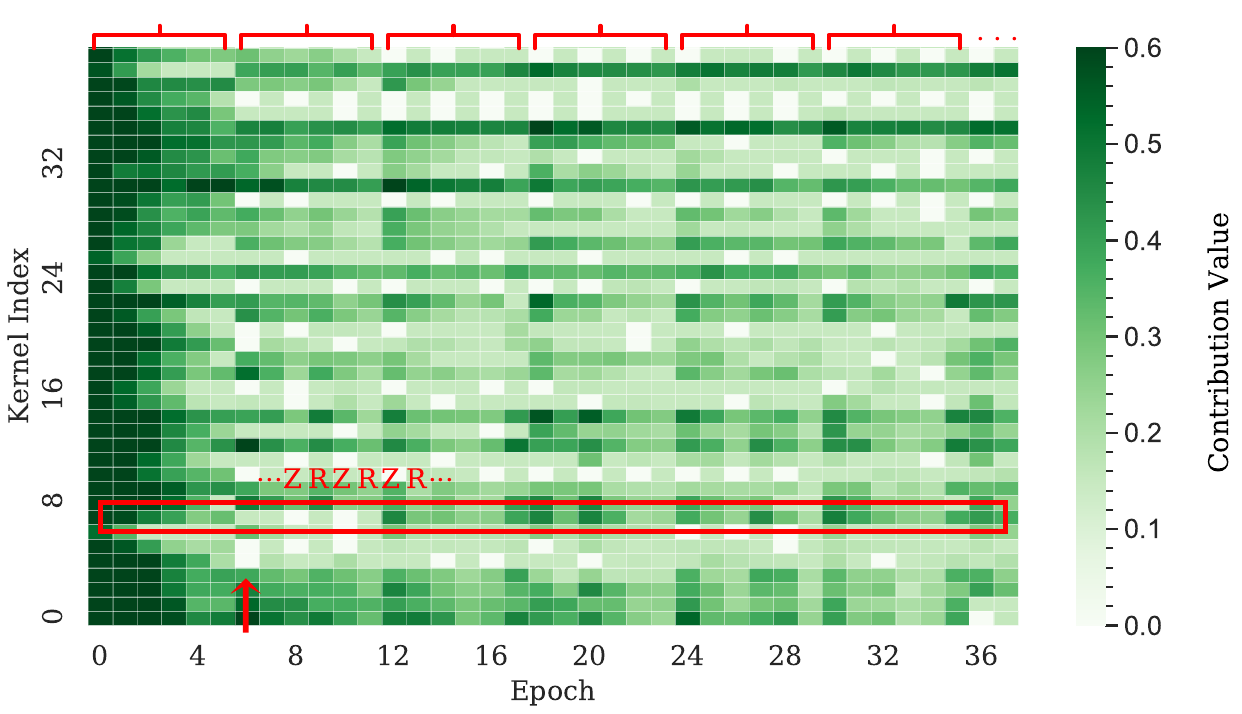}
    \caption{The contribution values of a local model. $\uparrow$ points to the communication round that starts zeroizing. \rotatebox{90}{\}} packages each local training.}
    \label{fig:zerobn_channel}
\end{figure}

Therefore, even though the current gradients of $\delta^{L}$ are zero, they can recover from zero by following the last updating directions. This process is fully automated and efficient, as no additional steps are required. These kernels zeroized in the previous zero epoch may become significant and we will not zeroize them in the future Zeroizing process. Thus, the architecture of the local model is changed, as illustrated in Fig. \ref{fig:zerobn_channel}. Moreover, the training process tries to minimize loss and improve accuracy, allowing the model architecture to improve for better accuracy. Our local learning gradually cultivates an optimal architecture for each system through the training process to maximize accuracy under the latency constraint.

\textbf{Training Overhead:} Compared to the local training of traditional FL, we add mask layers and a zeroizing process. As they are both related to the kernels in convolution layers, we mainly focus on the convolution. The time complexity of convolution layers for an input is $O(\sum_{l=1}^L M_l^2K_l^2C_{l-1}C_l)$, where $L$ is the number of convolution layers, $C_l$ is the number of the kernels in layer $l$, that is, the number of output channels of this layer, $M_l$ is the output size of layer $l$ and $K_l$ represents the kernel size of layer $l$. Mask layers have a time complexity of $O(\sum_{l=1}^L M_l^2C_l)$, which is much lower than convolution operations. The zeroizing operation includes a sorting step with a time complexity of $O(nlogn), n = \sum_{l=1}^LC_l$. Generally, $logn$ is much smaller than $M_l^2K_l^2C_{l-1}$ of each layer $l$, so its complexity is lower than the convolution layers. The latency predictor only includes hundreds of multiplication and addition operations, which is much smaller than convolution operations. Thus, the overhead of our local training can be neglected.





\subsection{Heterogeneous Aggregation -- Global Training}
\label{subsection:globaltraining}

FL maintains a global model from all locally trained models, combining all patterns learned from each private dataset. The parameters of local models are represented by \{ $W_l^1,W_l^2,...,W_l^n$ \}, where $l$ means different layers, and $n$ is the total number of local models. The global model receives all local model parameters and aggregates them into a global model $W_{gl}$. As shown in Fig.~\ref{fig:overview},  after several local training epochs, an aggregation process is conducted to combine all local parameters. Each local training - aggregation is a communication round, and it iterates multiple rounds. In the traditional identical FL algorithm, the aggregation is formulated as  $W_{gl}^{t+1} = \frac{1}{n} \sum_{i=1}^n (W_l^i)^{t+le}$ at the end of iteration $t$. At iteration $t+1$, $W_{gl}^{t+1}$ is transmitted to each local model and sets the local model parameters as  $(W_l^i)^{t+1} = W_{gl}^{t+1}$.
\begin{figure}[!t]
    \centering
    \includegraphics[trim={0pt 4pt 0pt 0pt},clip,width=0.48\textwidth]{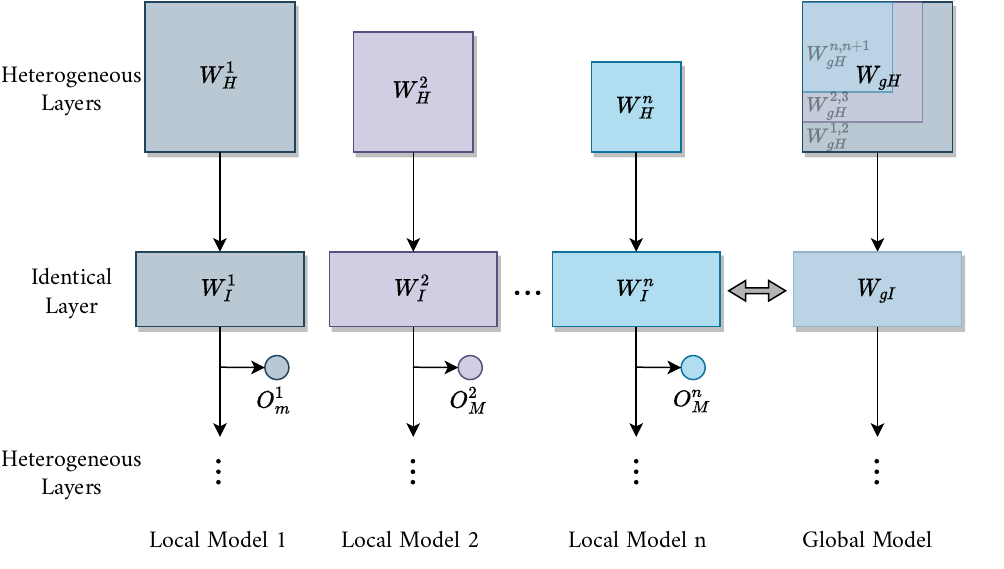} 
    \caption{Heterogeneous federated learning aggregation scheme of \ours.}
    \label{fig:fed}
\end{figure}

In this work, our primary motivation is to learn models for each system to meet its latency constraint while obtaining as much accuracy as possible. The details of our global training are shown in Algorithm~\ref{code:globaltraining}. We exploit model extension  (\textit{Line 4}) and shrinking  (\textit{Line 10}) to adjust each local model to suit its corresponding system, and these local models have a similar model architecture. Parameters of the smaller model are a subset of parameters of the larger model. Inspired by HeteroFL \cite{diao2020heterofl}, we first use an intuitive way to aggregate these heterogeneous local models (\textit{Line 13-16}), as shown in the rectangle in Fig.~\ref{fig:fed}. The model architecture mainly includes two parts, where the heterogeneous layers are shrunk or extended to change the latency, and the identical layer is for proto-correction.
The identical layer is a fully-connected layer, and this layer has the same shape among all local models. It can be aggregated following the traditional FL aggregation scheme as all local models share the same architecture. We focus more on the aggregation of parameters $W_H$ in heterogeneous layers. For notational convenience, we drop the iteration index $t$. Also, we aggregate parameters layer by layer, and we take one heterogeneous layer as an example. $W_H^i$ is the parameters of a heterogeneous layer of local model $i$, and $W_H^{i }\setminus W_H^{j}$ means the parameters in $W_H^i$ but not in $W_H^{j}$, $W_H^{i }\cap W_H^{j}$ represents the parameters in both $W_H^{i}$ and $W_H^{j}$, while  $W_H^{i }\cup W_H^{j}$ represents the parameters in $W_H^{i}$ or $W_H^{j}$.  $n$ is the total number of local models, and $n_{a}$ is the number of local models that come before local model $a$ in the complexity rank (\textit{Line 6}). Thus, the aggregation process can be represented in Eq.~(\ref{eq-fed-1})-(\ref{eq-fed-2}).

\begin{equation}
\begin{aligned}
\label{eq-fed-1}
    W_{gH}^{a, a+1} &= \frac{1}{n_{a+1}} \sum_{i=1}^{n} (W_H^{i }\setminus W_H^{a+1} \cap W_H^{a}), a < n \\
        W_{gH}^{n, n+1} &= \frac{1}{n} \sum_{i=1}^{n} (W_H^{i } \cap W_H^{n})
\end{aligned}
\end{equation}
\begin{equation}
\label{eq-fed-2}
    W_{gH} = W_{gH}^{1,2} \cup W_{gH}^{2,3} \cup \cdot \cdot \cdot \cup W_{gH}^{n,n+1}    
\end{equation}


\begin{algorithm}[t]
\label{code:globaltraining}
\KwIn{model, training settings: $ts$, local\_epoch: $le$, start\_zero round: $s$, latency constraint: $l$, client number: $n$.}
\KwOut{local models and their parameters}

\caption{FL Aggregation -- GlobalTrain}
\begin{algorithmic}[1] 
\STATE Initialize the original model and the prototype array $pt$;\
\STATE Duplicate the model for each client;\
\FOR{$m \leftarrow 1$ \textbf{to} $n$}
    \STATE Extend $model_m$ if its latency on $client_m$ less than $l$;\
\ENDFOR

\STATE Sort models by their complexity in descending order;\
\FOR{$c \leftarrow 1$ \textbf{to} $Round_{max}$}

    \FOR{$m \leftarrow 1$ \textbf{to} $n$ \textbf{in parallel}}
        \STATE Send $model_m$ to the corresponding $client_m$;\
        \STATE $model_m^{'}, pt_m \leftarrow $  LocalTrain($model_m, ts, c, s, le, l, pt$);\
    \ENDFOR
    \IF{$c \neq Round_{max}$}
            \STATE Derive each part of $W_{gH}$ by Eq.~(\ref{eq-fed-1}) and all $model_m^{'}$;\
        \FOR{$m \leftarrow 1$ \textbf{to} $n$}
            \STATE $model_m \leftarrow   \cup_{m}^{n} W_{gH}^{m,m+1}$;\ 
        \ENDFOR
    
        \STATE Merge all $pt_m$ and update $pt$ by Eq.~(\ref{eq-fed-3});\
    \ENDIF
\ENDFOR

\end{algorithmic}
\end{algorithm}

However, this method may lead to lower accuracy in a larger model architecture. Take Fig.~\ref{fig:fed} as an example, the parameters in $W_{gH}^{1,2}$ are only trained on private dataset $1$. These parameters impact the convergence of local mode $1$ severely, reducing its accuracy, particularly on the Non-IID data. Thus, we also proposed to use the prototype scheme \cite{xu2020attribute} to correct these parameters and enable these parameters to learn the pattern in other datasets  (\textit{Line 10, 17}).  As shown in the circle of Fig.~\ref{fig:fed}, $O_m^i$ is the output of the identical fully-connected layer near the final layer, and it can be used to represent a class by a high-dimension vector. This vector is called the prototype of this class. As shown in Eq.~(\ref{eq-fed-3}), we first aggregate all prototypes by class from all local models at training iteration $t$. Then, in the next training iteration $t+1$, we add a penalty item between the current prototype of a class and the prototype of this class from previous iterations into the loss function, as illustrated in Eq.~(\ref{eq-fed-4}), where $L_1$ represents the loss function of the model outputs to labels, $L_2$ means the loss function of the prototype and $\beta$ is a pre-defined coefficient. This is the proto-training used in Algorithm~\ref{code:localtraining}. In this way, larger models can perceive patterns of all private datasets to improve accuracy.

\begin{equation}
\label{eq-fed-3}
t:proto[l_j]^t =\frac{1}{n} \sum_{i=1}^{n} O_m^i[label == l_j], l_j \in label
\end{equation}
\begin{equation}
\label{eq-fed-4}
t+1:Loss = L_1(label, out) +\beta L_2(O_m, proto^t)
\end{equation}



\begin{figure*}

    \centering
            \setlength{\abovecaptionskip}{0pt}
	\subfigure{
                       \includegraphics[width=0.98\textwidth]{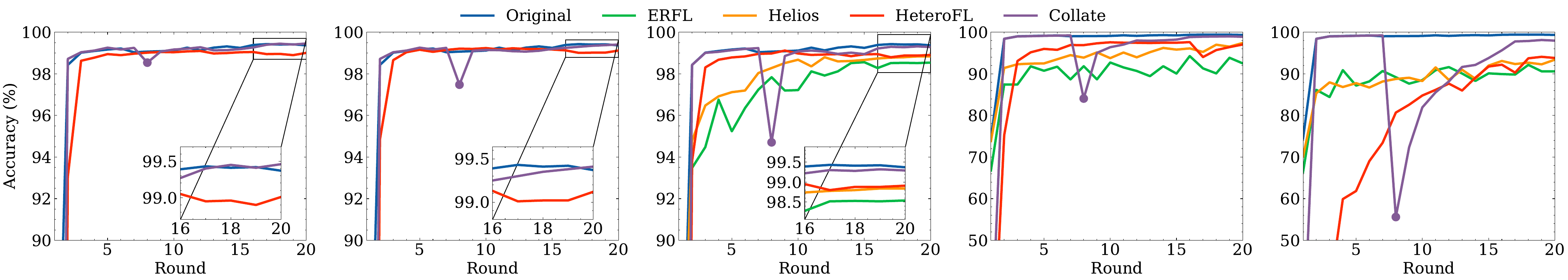}
	}\hspace{-8cm}
	\subfigure{
		\includegraphics[trim={0pt 4pt 0pt 32pt},clip,width=0.98\textwidth]{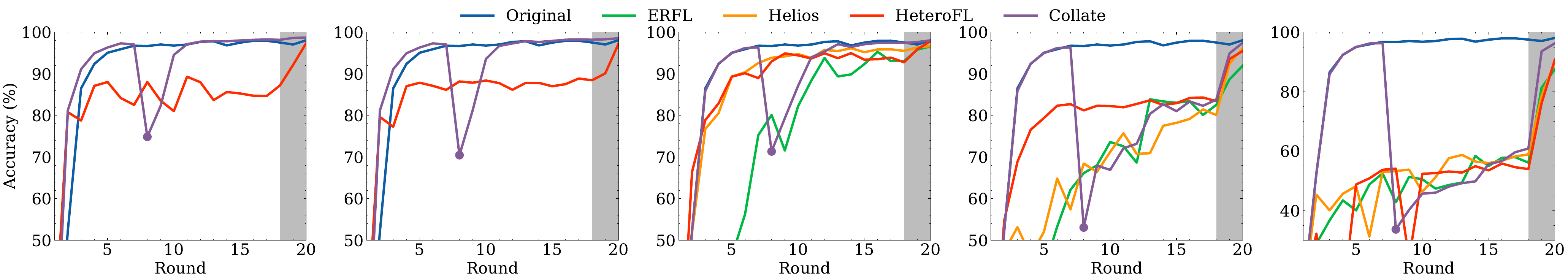} 

	}
	\caption{The test accuracy of LeNet on MNIST under different complexity levels  (From left to right are $1.5$, $1.25$, $1.0$, $0.5$, $0.25$). The top five show results on IID data and the bottom five are on Non-IID data. • points to the communication round that starts the zeroizing-recovering process of \ours.}
	\label{fig:mnist}
	   \vspace{-0.4cm}
\end{figure*}

\textbf{Training Overhead:} The global training adds the proto-corrected scheme. Derived from Eq.~(\ref{eq-fed-3})-(\ref{eq-fed-4}), the time complexity of this scheme is $O(IV)$, where $I$ is the total number of training data and $V$ is the dimension of $O_m^i$. This complexity is not comparable to a traditional aggregation scheme, which is only related to the number of model parameters and clients. But when considering the local training, its complexity is much lower than $O(I\sum_{l=1}^L M_l^2K_l^2C_{l-1}C_l)$. The model extension introduces overhead to the learning process, but it is intended to improve accuracy for powerful systems. And the model does not belong to the training algorithm, so the model extension is not included in the overhead of the training algorithm.

\section{Experiments}
\label{section:experiments}

In this section, we first present details of our implementation and the experimental environment, including hardware platforms, benchmark models and datasets. Then we discuss the performance of \ours, compared to state-of-the-art methods: Expanding the Reach of Federated Learning (ERFL) \cite{caldas2018expanding}, Helios \cite{xu2021helios} and HeteroFL \cite{diao2020heterofl}. Their brief is in Section \ref{section:relatedwork}.
\subsection{Experimental Setup}

\textbf{Hardware Platforms:} 
We collaboratively train models for five distinct devices to evaluate \ours. Each device and its inferred model constitute an edge system. These devices are HP ProBook 440 G6, NVIDIA Jetson TX2, NVIDIA Jetson Nano, Raspberry Pi 4B and Samsung Galaxy Note10. HP ProBook is equipped with an Intel Core i5-8265U CPU with 8GB of memory. TX2 is equipped with an NVIDIA Pascal GPU and 8GB of memory. Nano has an NVIDIA Maxwell GPU and 4GB of memory. Pi owns an ARM Cortex-A72 embedded CPU and 4GB of memory, and Note10 is equipped with a Qualcomm Snapdragon 855 processor and 8GB of memory. Also, \ours does not limit the number of clients.

\textbf{Datasets and Models:}
The datasets used for the benchmark are MNIST, CIFAR-10, CIFAR-100 and Human Activity Recognition (HAR) \cite{anguita2013public}, which are widely used in evaluating Federated Learning \cite{xu2021helios, diao2020heterofl, zhang2021parameterized, li2019fedmd, tan2022fedproto,caldas2018expanding, li2021hermes, jeong2018communication}. The first three are image classification datasets, which are popular computer vision applications. And HAR is an attractive feature for smartphones using data collected from different types of onboard sensors. We split the training data of each dataset into five parts, and each device owns a part and keeps it private. The trained model is evaluated on all test data of the corresponding dataset. For the generation of IID data, we randomly split the training data, and each class in a dataset exists on all five devices. And for the generation of Non-IID data, we force each class of a dataset to be divided into at most two parts, thus, a class only exists on at most two devices. The used models are LeNet, AlexNet, ResNet-18 and IMU \cite{moya2018convolutional}, and they are trained on MNIST, CIFAR-10, CIFAR-100 and HAR, respectively.

\textbf{Implementation Details:} The proposed method is implemented using the Pytorch library \cite{pytorch}. We set the local training epoch to $6$, and the communication rounds are $20$, $30$, $40$ and $100$ for LeNet, IMU, AlexNet and ResNet-18, respectively. And for all datasets, the optimizer is SGD with a momentum of $0.9$ and no weight decay. The learning rates are set to $0.05$,  $0.003$, $0.001$ and $0.01$ for LeNet, IMU, AlexNet and ResNet-18, respectively, which are widely used for these models. To exhibit the superiority of \ours under the latency constraint, we re-implement ERFL \cite{caldas2018expanding}, Helios \cite{xu2021helios} and HeteroFL \cite{diao2020heterofl} according to their papers or open-source codes. We use the latency predictor for each method and modify them to guarantee the latency constraint of each local model. And we use the same training settings for all methods.



\subsection{Accuracy Evaluation}
Our method can also learn models with pre-defined complexities, reflected by changing the $Shrink\_ratio$ to a constant in local training (\textit{Line 9} of Algorithm \ref{code:localtraining}). Due to HeteroFL \cite{diao2020heterofl} using pre-defined complexities to learn models, their method can easily include the model extension. We follow their work to set five discrete complexity levels of $1.5$, $1.25$, $1.0$, $0.5$, $0.25$ to evaluate \ours and HeteroFL \cite{diao2020heterofl}. However, ERFL \cite{caldas2018expanding} and Helios \cite{xu2021helios} can only shrink the model size, thus, we use $1.0$, $0.5$, $0.25$ complexity levels for these two works.
\begin{table}[t]
\centering
            \setlength{\abovecaptionskip}{1pt}
            \setlength{\belowcaptionskip}{-5pt} 
\caption{The test accuracy on HAR under different complexity levels.}
\label{tab:hariid}
   \setlength{\tabcolsep}{3pt} 
    \renewcommand{\arraystretch}{1.2} 
\begin{tabular}{ccccccc}
\hline
                                                                               &                                       & \multicolumn{5}{c}{\textbf{Accuracy (\%) of Different Complexities}}                                                                                                                                       \\ \cline{3-7} 
\multirow{-2}{*}{\textbf{\begin{tabular}[c]{@{}c@{}}Data\\ DIST\end{tabular}}} & \multirow{-2}{*}{\textbf{Method}}     & \textbf{C-1.50x}                       & \textbf{C-1.25x}                       & \textbf{C-1.00x}                       & \textbf{C-0.50x}                       & \textbf{C-0.25x}                       \\ \hline
                                                                               & \textbf{Original}                     & N.A.                                 & N.A.                                 & 94.53                                  & N.A.                                  & N.A.                                  \\
                                                                               & \textbf{ERFL \cite{caldas2018expanding}}                         & N.A.                                      & N.A.                                      & 93.87                                  & 89.64                                  & 81.53                                  \\
                                                                               & \textbf{Helios \cite{xu2021helios}}                       & N.A.                                      & N.A.                                      & 94.33                                  & 91.28                                  & 88.72                                  \\
                                                                               & \textbf{HeteroFL \cite{diao2020heterofl}}                     & 93.57                                  & 94.60                                  & \textbf{94.81}                         & 93.63                                  & 90.17                                  \\
\multirow{-5}{*}{\textbf{IID}}                                                 & \cellcolor[HTML]{DBDBDB}\textbf{\ours} & \cellcolor[HTML]{DBDBDB}\textbf{96.40} & \cellcolor[HTML]{DBDBDB}\textbf{96.33} & \cellcolor[HTML]{DBDBDB}94.40          & \cellcolor[HTML]{DBDBDB}\textbf{94.26} & \cellcolor[HTML]{DBDBDB}\textbf{91.68} \\ \hline
                                                                               & \textbf{Original}                     & N.A.             &N.A.             & 76.33              & N.A.              & N.A.              \\
                                                                               & \textbf{ERFL \cite{caldas2018expanding}}                         & N.A.                                      & N.A.                                      & 74.21                                  & 71.26                                  & 66.05                                  \\
                                                                               & \textbf{Helios \cite{xu2021helios}}                       & N.A.                                      & N.A.                                      & 75.16                                  & 71.63                                  & 67.87                                  \\
                                                                               & \textbf{HeteroFL \cite{diao2020heterofl}}                     & 75.03                                  & 75.70                                  & 75.67                                  & 73.68                                  & 69.32                                  \\
\multirow{-5}{*}{\textbf{Non-IID}}                                             & \cellcolor[HTML]{DBDBDB}\textbf{\ours} & \cellcolor[HTML]{DBDBDB}\textbf{77.54} & \cellcolor[HTML]{DBDBDB}\textbf{78.15} & \cellcolor[HTML]{DBDBDB}\textbf{76.38} & \cellcolor[HTML]{DBDBDB}\textbf{75.57} & \cellcolor[HTML]{DBDBDB}\textbf{71.53} \\ \hline
\end{tabular}
\end{table}

\textbf{Results on IID data:} The top five in Fig. \ref{fig:mnist} show that \ours achieves the highest accuracy compared to HeteroFL \cite{diao2020heterofl}, Helios \cite{xu2021helios}, and ERFL \cite{caldas2018expanding} under different complexity levels for the MNIST dataset. The original method is the traditional identical FL \cite{mcmahan2017communication}, and it obtains only one global model with the same complexity ($1.0$). We put its same curve in these five sub-figures for easy comparison. Compared with HeteroFL \cite{diao2020heterofl} on model extension, \ours with the proto-corrected scheme can achieve higher accuracy than the original method, while the accuracy of HeteroFL \cite{diao2020heterofl} is lower than the original method. Compared with Helios \cite{xu2021helios} and ERFL \cite{caldas2018expanding}, \ours can achieve the highest accuracy in the shrunk models through the well-designed Zeroizing-Recovering process. Table \ref{tab:hariid} (IID part) presents that on the HAR dataset, \ours can also perform well in most cases. All highest accuracy is marked in \textbf{boldface}. The accuracy of our extended models can outperform the original one by about $2\%$ and our shrunk models achieve the lowest accuracy loss.

\textbf{Results on Non-IID data:} The bottom five in Fig. \ref{fig:mnist} demonstrate that \ours still performs well on the Non-IID data of the MNIST dataset. Different from IID data, for each method and each dataset, we use the last $10\%$ of the total communication rounds to fine-tune each local model in a traditional FL way \cite{mcmahan2017communication}. This is because final models are generated after local training, and aggregation updates zeroized weights into non-zero in all methods (except HeretoFL, but for fairness, we also fine-tune HeteroFL). Updating only by the local dataset impacts its accuracy on all test sets, especially under Non-IID data. Fig. \ref{fig:mnist} also shows that some of our models perform well without fine-tuning by the proto-corrected scheme. Table \ref{tab:hariid} (Non-IID part) shows that on the HAR dataset, \ours can still achieve the highest accuracy.

\begin{table}[t]
            \setlength{\abovecaptionskip}{1pt}
            \setlength{\belowcaptionskip}{-5pt} 
\centering
\caption{The test accuracy on Cifar-10 under the latency constraint.}
\label{tab:cifariid}
   \setlength{\tabcolsep}{3pt} 
    \renewcommand{\arraystretch}{1.2} 

\begin{tabular}{ccccccc}
\hline
                                                                               &                                       & \multicolumn{5}{c}{\textbf{Accuracy (\%) of Local Models}}                                                                                                                                                         \\ \cline{3-7} 
\multirow{-2}{*}{\textbf{\begin{tabular}[c]{@{}c@{}}Data\\ DIST\end{tabular}}} & \multirow{-2}{*}{\textbf{Method}}     & \textbf{ProBook}                           & \textbf{Tx2}                               & \textbf{Nano}                          & \textbf{Pi}                            & \textbf{Note10}                        \\ \hline
                                                                               & \textbf{Original}                     & 82.72                                    & 82.72                                      & \textbf{82.72}                                  & N.A.                                  & N.A.                                 \\
                                                                               & \textbf{ERFL \cite{caldas2018expanding}}                         & 76.53                                      & 76.53                                      & 76.53                                  & 75.80                                  & 76.28                                  \\
                                                                               & \textbf{Helios \cite{xu2021helios}}                       & 81.24                                      & 81.24                                      & 81.24                                  & 79.30                                  & 79.42                                  \\
                                                                               & \textbf{HeteroFL \cite{diao2020heterofl}}                     & 82.20 (E)                                  & 81.11 (E)                                  & 82.23                         & 74.35                                  & 78.66                                  \\
\multirow{-5}{*}{\textbf{IID}}                                                 & \cellcolor[HTML]{DBDBDB}\textbf{\ours} & \cellcolor[HTML]{DBDBDB}\textbf{83.74 (E)} & \cellcolor[HTML]{DBDBDB}\textbf{83.93 (E)} & \cellcolor[HTML]{DBDBDB}81.82          & \cellcolor[HTML]{DBDBDB}\textbf{80.40}  & \cellcolor[HTML]{DBDBDB}\textbf{81.72} \\ \hline
                                                                               & \textbf{Original}                     & 80.25                                      & 80.25                                      & 80.25                                  & N.A.                                & N.A.                                  \\
                                                                               & \textbf{ERFL \cite{caldas2018expanding}}                         & 75.80                                      & 75.80                                      & 75.80                                  & 63.06                                  & 75.13                                  \\
                                                                               & \textbf{Helios \cite{xu2021helios}}                       & 79.82                                      & 79.82                                      & 79.82                                  & 69.62                                  & 76.28                                  \\
                                                                               & \textbf{HeteroFL \cite{diao2020heterofl}}                     & 80.15 (E)                                  & 79.02 (E)                                  & 80.13                                  & 69.15                                  & 76.86                                  \\
\multirow{-5}{*}{\textbf{Non-IID}}                                             & \cellcolor[HTML]{DBDBDB}\textbf{\ours} & \cellcolor[HTML]{DBDBDB}\textbf{81.45 (E)} & \cellcolor[HTML]{DBDBDB}\textbf{81.11 (E)} & \cellcolor[HTML]{DBDBDB}\textbf{80.66} & \cellcolor[HTML]{DBDBDB}\textbf{74.83} & \cellcolor[HTML]{DBDBDB}\textbf{78.29} \\ \hline
\end{tabular}
\end{table}

\subsection{Latency-Accuracy Evaluation}

To demonstrate the effectiveness of our method for multiple latency-critical edge systems, we evaluate the accuracy of \ours and state-of-the-art heterogeneous FL methods under a pre-defined latency constraint. As Helios used Nvidia Jetson Nano as the non-straggler \cite{xu2021helios}, we also pick the original model inference latency on Jetson Nano as the constraint, which means we adjust all local models to achieve a similar latency as that of the original corresponding model on Jetson Nano.

Table \ref{tab:cifariid} presents that compared with ERFL \cite{caldas2018expanding}, Helios \cite{xu2021helios} and HeteroFL \cite{diao2020heterofl}, \ours can achieve the best accuracy for almost all the five platforms on IID or Non-IID data of the CIFAR-10 dataset. As HeteroFL \cite{diao2020heterofl} and our method can extend models to larger than the original ones, while others can only shrink models, we use $E$-flag to represent model extension. Additionally, each local model in Table \ref{tab:cifariid} can meet the latency constraint ($38$ ms) and achieve a latency in the range of $36.8\pm 0.7 ms$. For methods without extension, the latency of their models is lower than the latency constraint on powerful devices.
Compared to the original model, for the IID data, we extend the model on HP ProBook from about $23.0$ ms and improve the accuracy by $1.02\%$. Also, the latency of the original model on Note10 is around $82.6$ ms, and we reduce its latency by about $54.0\%$ with only a $1.00\%$ accuracy drop. For the Non-IID data, since it is more difficult to learn the pattern of the dataset, the benefit of \ours is less than that of IID data, but our method still outperforms other methods.

Table \ref{tab:cifar100iid} shows the evaluation results of \ours on the CIFAR-100 dataset. Each local model in Table \ref{tab:cifar100iid} is under a latency constraint of about $70$ ms and they achieve a latency in the range of $67.8\pm 1.8 ms$. Among these models, \ours obtains the best accuracy under all conditions. Compared to the original model, we extend it on Jetson TX2 from about $27.4$ ms and obtain a $2.02\%$ accuracy improvement. Also, the latency of the original model on Pi is around $152.6$ ms, and we reduce its latency by about $54.1\%$ with only a $1.71\%$ accuracy drop.  Specifically, from Table \ref{tab:cifariid}-\ref{tab:cifar100iid}, we can conclude that, compared to other methods \cite{xu2021helios, diao2020heterofl, caldas2018expanding}, \ours improves the accuracy by $1.96\%$ on average for extended models, and our shrunk models can obtain a $3.09\%$ accuracy improvement on average.



\section{Conclusion}
\label{section:conclusion}
In this work, we have presented a collaborative neural network learning framework for multiple latency-critical edge systems to directly satisfy the latency constraint of DNNs on their devices. The collaboration augmented the training data for each system to improve accuracy while also protecting their data privacy. Our framework can satisfy the latency constraints of each system with one training process. It contained two components: a dynamic zeroizing-recovering algorithm and a proto-corrected aggregation scheme. The dynamic algorithm can extract optimal model architectures to achieve less accuracy drop under a large latency reduction. Our aggregation scheme improved the accuracy of extended models. We also included the latency predictor and model extension to guide and optimize this framework. Experiments verified that our framework has obtained remarkable improvements.



\begin{table}[]
\centering
\caption{The test accuracy on Cifar-100 under the latency constraint.}
\label{tab:cifar100iid}
   \setlength{\tabcolsep}{3pt} 
    \renewcommand{\arraystretch}{1.2} 

\begin{tabular}{ccccccc}
\hline
                                                                               &                                       & \multicolumn{5}{c}{\textbf{Accuracy (\%) of Local Models}}                                                                                                                                                         \\ \cline{3-7} 
\multirow{-2}{*}{\textbf{\begin{tabular}[c]{@{}c@{}}Data\\ DIST\end{tabular}}} & \multirow{-2}{*}{\textbf{Method}}     & \textbf{ProBook}                           & \textbf{Tx2}                               & \textbf{Nano}                          & \textbf{Pi}                            & \textbf{Note10}                        \\ \hline
                                                                               & \textbf{Original}                     & 70.25                                      & 70.25                                      & 70.25                                  & N.A.                                  & N.A.                                  \\
                                                                               & \textbf{ERFL \cite{caldas2018expanding}}                         & 69.78                                      & 69.78                                      & 69.78                                  & 68.30                                  & 69.82                                  \\
                                                                               & \textbf{Helios \cite{xu2021helios}}                       & 70.14                                      & 70.14                                      & 70.14                                  & 68.30                                  & 70.23                                  \\
                                                                               & \textbf{HeteroFL \cite{diao2020heterofl}}                     & 68.70 (E)                                  & 70.25 (E)                                  & 70.35                                  & 67.25                                  & 70.48                                  \\
\multirow{-5}{*}{\textbf{IID}}                                                 & \cellcolor[HTML]{DBDBDB}\textbf{\ours} & \cellcolor[HTML]{DBDBDB}\textbf{70.68 (E)} & \cellcolor[HTML]{DBDBDB}\textbf{72.27 (E)} & \cellcolor[HTML]{DBDBDB}\textbf{70.69} & \cellcolor[HTML]{DBDBDB}\textbf{68.54} & \cellcolor[HTML]{DBDBDB}\textbf{70.87} \\ \hline
\end{tabular}
\end{table}


\bibliographystyle{IEEEtran}
\bibliography{reference}

\end{document}